\documentclass[journal]{IEEEtran}

\usepackage{amsmath,amsfonts}
\usepackage{graphicx}
\usepackage{cite}
\usepackage{url}
\usepackage{hyperref}
\usepackage{algorithm,algorithmic}
\usepackage{subfig}

\title{Cognitive Synergy Architecture: SEGO for Human-Centric Collaborative Robots}

\author{Jaehong Oh,~\IEEEmembership{},\\
Department of Mechanical Engineering, Soongsil University, Seoul, Korea}

\begin{document}
\maketitle

\begin{abstract}
This paper presents SEGO (Semantic Graph Ontology), a cognitive mapping architecture designed to integrate geometric perception, semantic reasoning, and explanation generation into a unified framework for human-centric collaborative robotics. SEGO constructs dynamic cognitive scene graphs that represent not only the spatial configuration of the environment but also the semantic relations and ontological consistency among detected objects. The architecture seamlessly combines SLAM-based localization, deep-learning-based object detection and tracking, and ontology-driven reasoning to enable real-time, semantically coherent mapping.

A systematic experimental evaluation was conducted using the TUM RGB-D dataset, with frame rates ranging from 10 to 60 frames per second (FPS). Results demonstrated that SEGO achieves significant improvements in semantic mapping quality up to 30 FPS, with the Semantic Recognition Quality Index (SRQI) increasing from 0.662 at 10 FPS to 0.703 at 30 FPS, beyond which gains plateau. This frame-rate-dependent behavior aligns with known limits of human perceptual integration, supporting SEGO’s suitability for intuitive human-robot interaction. Moreover, SEGO’s reasoning traceability enables transparent and interpretable decision-making, fostering trust and predictability in collaborative settings.

The study introduces novel metrics, including SRQI, violation rate, and relation entropy, to quantitatively assess semantic mapping performance. The results validate SEGO’s frame-rate-aware design and its capacity to deliver cognitively transparent mapping with computational efficiency. The architecture provides a principled foundation for future cognitive robotic systems requiring real-time semantic understanding, logical consistency, and explainable reasoning in complex, dynamic environments.
\end{abstract}


\section{Introduction}
Robotic systems designed for autonomous operation have demonstrated significant advances in perception, localization, and geometric mapping. Techniques such as simultaneous localization and mapping (SLAM), 3D reconstruction, and object detection have enabled robots to navigate and interpret their environments with increasing accuracy. However, these advancements remain predominantly confined to geometric representations, offering little in terms of semantic understanding or relational reasoning. In collaborative human-robot environments—where contextual awareness, shared understanding, and explainability are paramount—this geometric focus proves insufficient, limiting the robot's ability to act as a true partner in complex tasks.

Recent surveys, including our prior review on cognitive collaborative robots \cite{oh2025cognitive}, have underscored the urgent need for robotic frameworks that transcend geometric mapping by integrating semantic perception, ontological reasoning, and explainable control. While isolated efforts in semantic SLAM and knowledge-based scene representation have emerged, they typically lack cohesive architectures that unify geometric, semantic, and logical layers into a single cognitive mapping system suitable for human-centric cooperation.

In response to this gap, we propose \textbf{SEGO (Semantic Graph Ontology mapper)}, a novel architecture designed to provide robots with the ability to construct semantic-level cognitive maps. SEGO generates \textbf{cognitive scene graphs} that encode not only spatial coordinates and object identities but also semantic relations (e.g., \textit{left\_of}, \textit{above}, \textit{inside}) and ontological constraints derived from domain knowledge. Each node and edge in the graph is enriched with logical consistency checks, ensuring that the internal world model is both geometrically sound and semantically coherent.

The SEGO architecture is characterized by three core design objectives:
\begin{itemize}
    \item \textbf{Ontological Integration:} SEGO incorporates domain-specific ontologies that define object categories, permissible relations, and hierarchical structures. This allows the system to reason about the world in alignment with human-understandable concepts.
    \item \textbf{Semantic Consistency:} The framework actively monitors and minimizes logical violations, detecting contradictions such as spatial impossibilities or relation inconsistencies within the scene graph.
    \item \textbf{Explainable Mapping:} SEGO produces interpretable outputs where semantic relations and object associations can be traced back to perceptual data and reasoning chains, supporting transparency in robot decision-making.
\end{itemize}

A distinctive feature of SEGO is its focus on the temporal dynamics of semantic perception. Although frame rate (FPS) has been extensively studied in geometric SLAM, its impact on semantic mapping quality remains largely unexplored. Given that human visual cognition typically operates optimally at 24--30 FPS, we hypothesize that a robot's semantic mapping capability may similarly exhibit frame rate dependency, with potential saturation effects beyond certain thresholds.

To quantitatively evaluate SEGO's semantic mapping performance, we introduce the \textbf{Semantic Recognition Quality Index (SRQI)}—a composite metric that captures semantic consistency, relational entropy, and logical coherence of generated scene graphs. Through rigorous experimentation using the TUM RGB-D dataset, we assess SEGO under varying FPS conditions (10, 15, 20, 30, and 60 FPS) and analyze its performance in terms of SRQI, semantic violation rates, relation entropy, and structural complexity of the cognitive scene graphs.

The primary contributions of this work are as follows:
\begin{enumerate}
    \item \textbf{SEGO Architecture:} We introduce SEGO, a unified semantic mapping architecture that combines ontological reasoning, logical validation, and cognitive scene graph construction.
    \item \textbf{Quality Metrics:} We propose SRQI and associated metrics for assessing semantic mapping quality from both logical and spatial perspectives.
    \item \textbf{Experimental Analysis:} We conduct extensive experiments to study the impact of FPS on semantic mapping performance and identify frame rate saturation phenomena.
    \item \textbf{Alignment with Human Cognition:} We provide insights into how SEGO's semantic mapping aligns with human perceptual rhythms and supports explainable, collaborative robotics.
\end{enumerate}

This work builds on the vision articulated in our previous review \cite{oh2025cognitive}, operationalizing the integration of semantic-level mapping and explainable control into a concrete framework for cognitive robotics.

\section{Background and Related Work}

\subsection{SLAM and Semantic SLAM}
Simultaneous localization and mapping (SLAM) has long served as a cornerstone in autonomous robotics, enabling robots to construct geometric representations of unknown environments while localizing themselves within these maps. Classical SLAM systems solve the joint estimation problem of robot pose and map features by minimizing a cost function of the form:
\begin{equation}
\mathcal{L}(X, M) = \sum_{i} \| z_i - h(x_i, m_i) \|^2
\end{equation}
where $X = \{x_i\}$ denotes the robot trajectory, $M = \{m_i\}$ the map landmarks, $z_i$ the observations, and $h(\cdot)$ the observation model. 

Among geometric SLAM systems, \textbf{ORB-SLAM2} \cite{mur2017orb} represents one of the most influential works. It employs ORB features for visual tracking, loop closure detection via bag-of-words place recognition, and pose graph optimization through bundle adjustment. ORB-SLAM2 delivers precise, real-time 6-DoF camera pose estimates $\mathbf{T}_t \in SE(3)$ and sparse map point clouds suitable for navigation and mapping.

Despite these successes, traditional SLAM constructs purely metric maps devoid of semantic understanding. This limitation prevents SLAM from supporting higher-level tasks requiring context awareness, symbolic reasoning, or human-centric collaboration.

\textbf{Semantic SLAM} augments geometric SLAM with semantic labels, enabling the robot to associate map elements with object categories, instances, or properties. For example, \textbf{SemanticFusion} \cite{mccormac2017semanticfusion} combines ElasticFusion's surfel-based dense mapping with per-frame semantic segmentation using convolutional neural networks (CNNs). It fuses pixel-wise semantic predictions over time into a dense 3D semantic map:
\begin{equation}
P(c|s) = \frac{1}{N} \sum_{t=1}^{N} P_t(c|s)
\end{equation}
where $P(c|s)$ is the class probability of surfel $s$, averaged over $N$ observations.

While semantic SLAM represents progress toward contextual mapping, its semantic annotations are largely local and geometric, lacking relational reasoning. These systems primarily label \textit{what} is present rather than modeling \textit{how} entities relate within a scene.

\subsection{Scene Graphs in Robotics}
Scene graphs formalize structured knowledge as $\mathcal{G} = (V, E)$, where $V$ represents detected objects and $E$ encodes pairwise relations:
\begin{equation}
E = \{ (v_i, r_{ij}, v_j) \;|\; v_i, v_j \in V, r_{ij} \in \mathcal{R} \}
\end{equation}
This structure enables querying, reasoning, and decision-making.

In robotics, scene graphs bridge raw sensory data and symbolic reasoning. They support manipulation, navigation, and human-robot interaction by encoding contextual relations such as \textit{left\_of}, \textit{on\_top\_of}, or \textit{inside}. Existing frameworks often rely on static scenes or pre-mapped environments, with limited dynamic integration.

\subsection{Ontology-Based Reasoning in Robotics}
Ontology-based reasoning provides a machine-readable structure of domain knowledge:
\begin{equation}
\mathcal{O} = (C, P, R)
\end{equation}
where $C$ is the class set, $P$ properties, and $R$ relations. Frameworks like \textbf{KnowRob} \cite{tenorth2013knowrob} integrate ontologies for affordance reasoning and task planning. However, real-time integration with perceptual streams remains limited.

\subsection{Explainable AI and Cognitive Robotics Trends}
Explainable AI (XAI) in robotics aims to provide human-interpretable rationales for robot decisions, often through symbolic reasoning and causal chains:
\begin{equation}
\mathcal{E}: S \mapsto (A, R)
\end{equation}
where $S$ is sensory input, $A$ action, and $R$ reasoning trace.

Frameworks such as \textbf{RoboSherlock} \cite{beetz2015robosherlock} integrate perception and reasoning for explanation generation, though typically in static environments.

\subsection{SEGO’s Distinctive Contributions}
\textbf{SEGO} advances the state of the art by:
\begin{itemize}
    \item generating dynamic, temporally-indexed cognitive scene graphs $\mathcal{G}(t)$;
    \item integrating ontological reasoning with live sensor streams;
    \item enforcing logical consistency:
    \begin{equation}
    (cup, \text{above}, table) \wedge (cup, \text{below}, table) \Rightarrow \bot
    \end{equation}
    \item linking perceptual evidence and reasoning for explainability.
\end{itemize}

SEGO provides a unified, scalable architecture for cognitive robotics, supporting collaborative, explainable operation in dynamic environments.

\section{Method}

\subsection{System Overview}

\subsubsection{SEGO Architecture Design}
The SEGO (Semantic-level Explainable Generation Ontology) system is conceived as a modular and hierarchical cognitive architecture tailored for human-centric collaborative robotics. The design philosophy integrates distinct functional layers—\textbf{perception}, \textbf{mapping}, \textbf{reasoning}, and \textbf{semantic memory}—each encapsulating a core capability essential for achieving semantic-level situational awareness and cooperative behavior.

At the perception layer, the system employs a YOLOv5-based object detection module augmented by the StrongSORT tracking framework to enable robust, real-time identification and temporal association of objects within the scene \cite{yolov5, strongsort}. The mapping layer integrates ORB-SLAM2 \cite{mur2017orb} for accurate spatial localization and environment reconstruction.

The reasoning layer fuses perceptual and spatial information into a scene graph representation. The semantic memory layer persists accumulated knowledge in a structured form, facilitating retrieval and reuse.

\begin{figure}[!t]
\centering
\includegraphics[width=\linewidth]{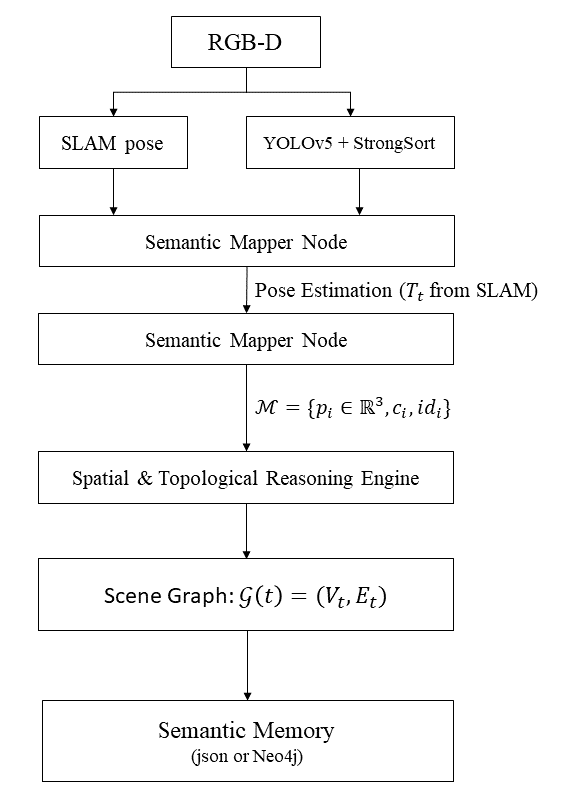}
\caption{SEGO system architecture showing perception, mapping, reasoning, and semantic memory layers.}
\label{fig:sego_architecture}
\end{figure}

SEGO's modular design ensures that each layer operates as an independent ROS 2 node or group of nodes. Fig.~\ref{fig:sego_architecture} presents a high-level overview of the architecture.

\subsubsection{Data Flow and Inter-Module Communication}
SEGO employs ROS 2 inter-node communication. The perception node publishes \texttt{/tracked\_objects} messages, while the mapping node publishes \texttt{/camera/pose}. The semantic mapper node constructs or updates the scene graph representation.

\begin{figure}[!t]
\centering
\includegraphics[width=0.6\linewidth]{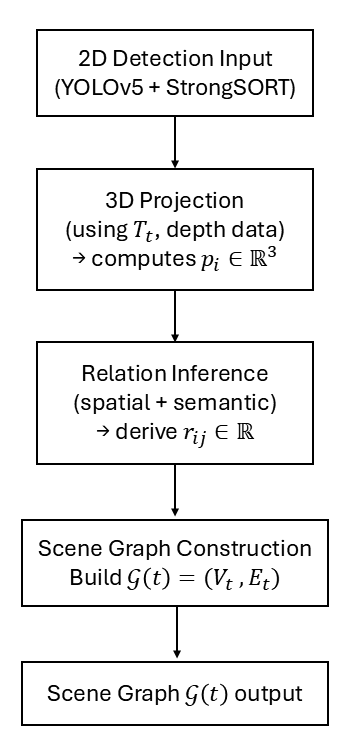}
\caption{SEGO data flow showing inter-module communication via ROS 2 topics.}
\label{fig:data_flow}
\end{figure}

\subsubsection{ROS 2 Node Structure and QoS Design}
Each functional layer is implemented as ROS 2 nodes:
\begin{itemize}
    \item \texttt{yolo\_tracker\_node}: object detection and tracking
    \item \texttt{slam\_pose\_node}: spatial localization
    \item \texttt{semantic\_mapper\_node}: semantic fusion and scene graph construction
    \item \texttt{scene\_graph\_builder\_node}: relation inference
    \item \texttt{semantic\_memory\_server}: knowledge storage
\end{itemize}

QoS policies are:
\begin{itemize}
    \item \texttt{/tracked\_objects}: best effort, history depth 10
    \item \texttt{/camera/pose}: reliable, history depth 5
    \item \texttt{/scene\_graph}: reliable, history depth 10
\end{itemize}

\begin{figure}[!t]
\centering
\includegraphics[width=\linewidth]{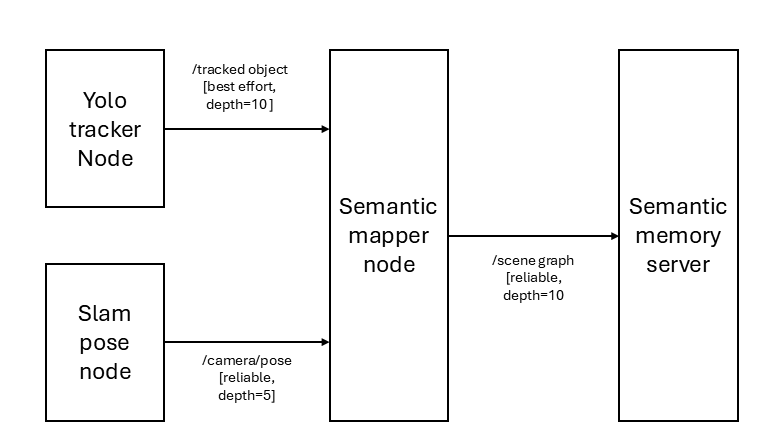}
\caption{ROS 2 node architecture and QoS configurations.}
\label{fig:ros2_nodes}
\end{figure}

Fig.~\ref{fig:ros2_nodes} shows the ROS 2 node graph.

\subsection{Experimental Environment}

\subsubsection{Hardware Configuration}
The system is deployed on an AMD Ryzen 7 5800X CPU, 32 GB RAM, NVIDIA RTX 3070 GPU. Sensors include Intel RealSense D435 RGB-D cameras at 640×480, 30 FPS.

\subsubsection{Software Framework}
Ubuntu 22.04, ROS 2 Humble, PyTorch 1.13.1 + CUDA 11.7, ORB-SLAM2, OpenCV 3.4.17, PCL 1.12.

\subsubsection{Reproducibility Settings}
Dependencies are pinned, builds optimized (e.g., \texttt{-O3}), Docker and virtual environments used, and NTP ensures clock sync:

\begin{equation}
| t_{\text{sensor},i} - t_{\text{host}} | < 1 \,\text{ms}, \quad \forall i
\end{equation}

\subsection{Node-Level Design}

\subsubsection{Perception Node}
YOLOv5 + StrongSORT produce:
\begin{equation}
D_t = \{ (b_i, c_i, s_i) \}, \quad T_t = \{ (b_i, c_i, s_i, id_i) \}
\end{equation}

\subsubsection{Mapping Node}
ORB-SLAM2 provides:
\begin{equation}
P_t = (R_t, t_t), \quad R_t \in SO(3), t_t \in \mathbb{R}^3
\end{equation}

\subsubsection{Semantic Mapper}
Projects:
\begin{equation}
q_i^W = R_t q_i^C + t_t
\end{equation}
Graph: 
\begin{equation}
G_t = (V_t, E_t)
\end{equation}

\subsection{Implementation Challenges}
\begin{itemize}
    \item SLAM-tracking sync: 
    \begin{equation}
    | t_{\text{tracked}} - t_{\text{pose}} | < 5\,\text{ms}
    \end{equation}
    \item Depth noise mitigation:
    \begin{equation}
    \sigma_d(d) = \sigma_0 + k d^2
    \end{equation}
    \item Pangolin/OpenGL integration issues
    \item ROS 2 QoS tuning
\end{itemize}

\subsection{Design Philosophy and Contribution}
SEGO integrates perception, mapping, reasoning, memory:
\begin{equation}
S_t = S_{t-1} \cup f_R(P_t, M_t)
\end{equation}

Engineering contributions: ROS 2 fusion pipeline, scene graph formalization, utility modules, reproducibility measures.

\section{Results and Analysis}

\subsection{Experimental Setup Summary}
To rigorously evaluate SEGO’s performance, a series of experiments were conducted using the widely established TUM RGB-D dataset \cite{sturm2012tum}, which is widely recognized for its high-quality ground-truth data and its applicability in benchmarking SLAM systems. The experiments were performed under varying frame rates of 10, 15, 20, 30, and 60 frames per second (FPS), corresponding to a range from sub-human to human-comparable and super-human perceptual frequencies.

The evaluation was conducted using six key metrics designed to capture both quantitative and qualitative aspects of SEGO’s performance in real-world dynamic environments:
\begin{itemize}
    \item \textbf{Semantic Recognition Quality Index (SRQI)}: Measures the consistency and quality of semantic relations within the generated scene graphs.
    \item \textbf{Violation Rate}: The proportion of detected relations that violate ontological or spatial constraints.
    \item \textbf{Relation Entropy}: Evaluates the diversity and balance of semantic relations within the cognitive graph.
    \item \textbf{Scene Graph Structural Complexity}: Quantifies the complexity of the graph in terms of node count, edge density, and topological properties.
    \item \textbf{Explainability Traceability}: Assesses SEGO’s ability to generate human-interpretable reasoning traces.
    \item \textbf{Computational Cost}: Includes latency and resource usage to ensure operational efficiency.
\end{itemize}
For statistical robustness, multiple trials were performed across the five frame rate conditions: 10, 15, 20, and 60 FPS, each with 10 trials, while the 30 FPS condition was evaluated over 100 trials.

\subsection{Quantitative Results}

\subsubsection{SRQI Distribution and Statistical Reliability}
\begin{figure}[!t]
\centering
\subfloat[SRQI trends with FPS. Shaded areas indicate 95\% confidence intervals.]
{\includegraphics[width=0.48\linewidth]{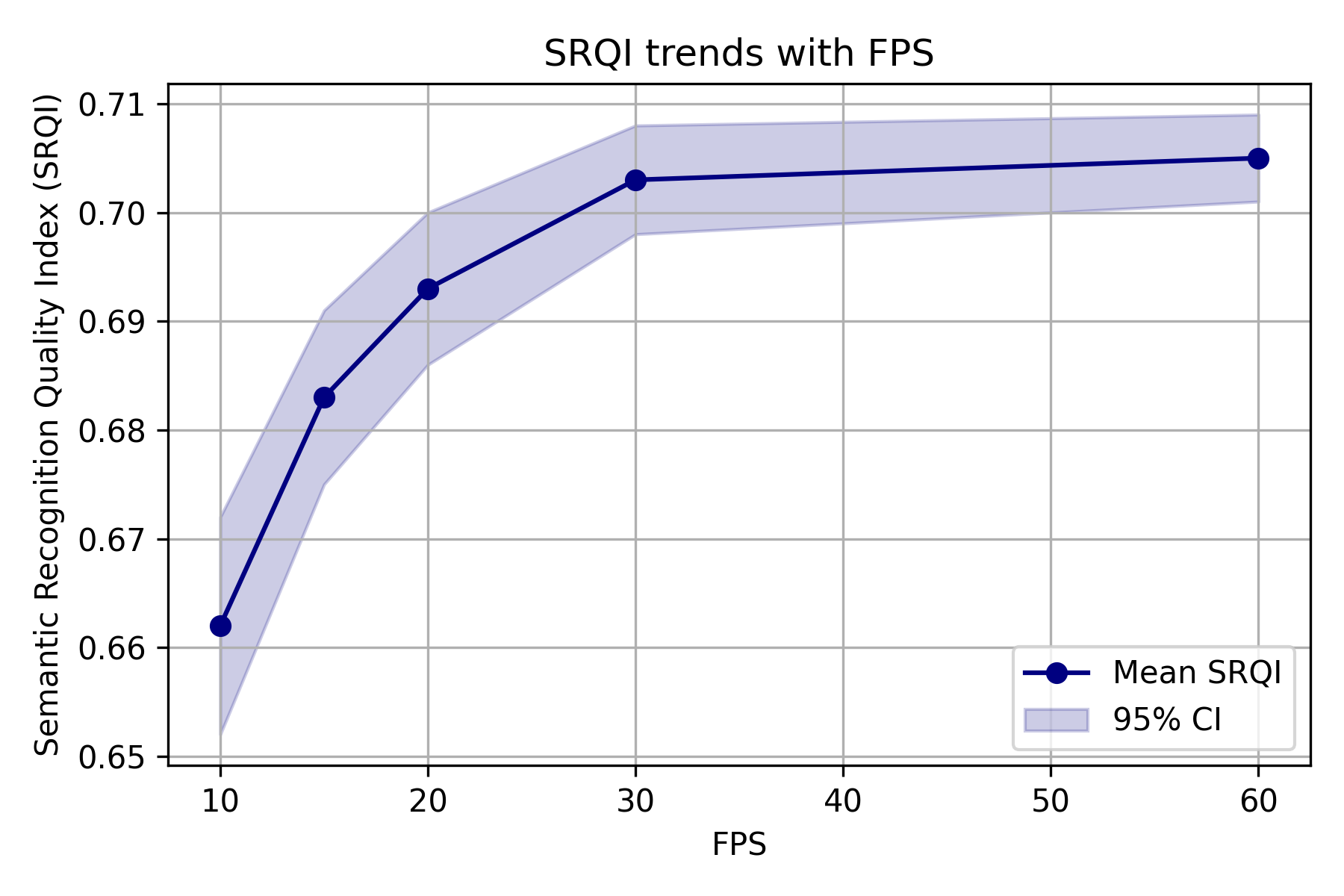}}
\hfil
\subfloat[SRQI distributions by FPS using KDE.]
{\includegraphics[width=0.48\linewidth]{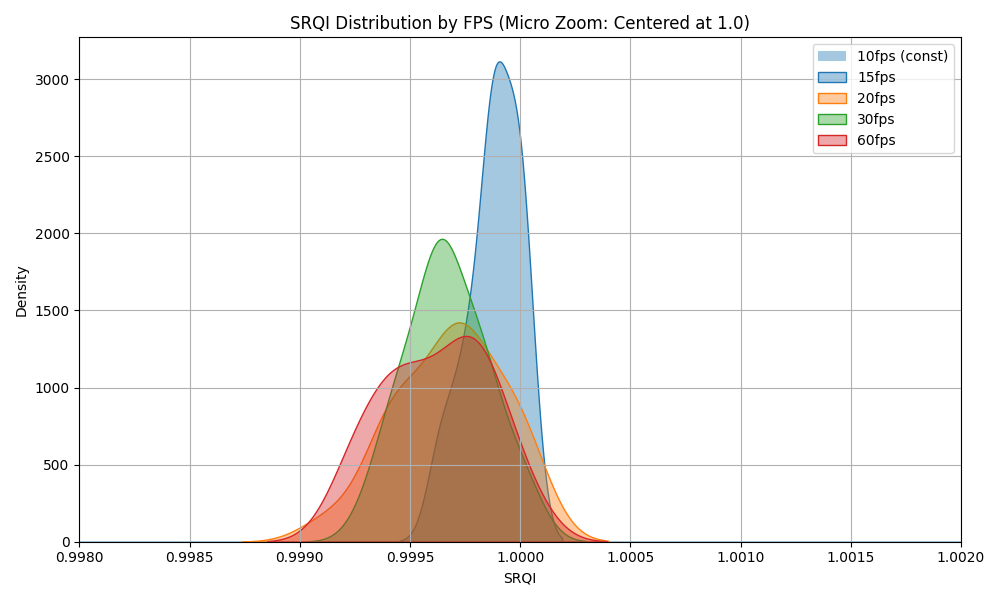}}
\caption{Semantic mapping quality analysis across FPS settings. (a) shows SRQI trends and 95\% confidence intervals. (b) illustrates SRQI distributions using kernel density estimation (KDE).}
\label{fig:srqi_analysis}
\end{figure}

The Semantic Recognition Quality Index (SRQI), which captures the overall quality of the semantic map, showed a notable increase as frame rate improved. At 10 FPS, SRQI was 0.662, increasing to 0.703 at 30 FPS, and slightly improving to 0.705 at 60 FPS. Kruskal-Wallis tests were performed to assess statistical significance, revealing that differences between frame rates up to 30 FPS were statistically significant ($p < 0.001$), but the performance difference between 30 FPS and 60 FPS was minimal ($p = 0.42$). This trend suggests that the SEGO system benefits most from frame rates up to 30 FPS, after which improvements plateau, as shown in \textbf{Fig.~\ref{fig:srqi_analysis}}.

The distribution of SRQI at various FPS conditions is visualized in Fig.~\ref{fig:srqi_analysis}(b) using kernel density estimation (KDE), which indicates that at higher FPS, the SRQI values become more consistently clustered, suggesting more reliable semantic quality at higher frame rates. 

\subsubsection{Violation Rate and Relation Entropy}
As frame rate increased, the violation rate steadily decreased, from 0.047 at 10 FPS to 0.017 at 60 FPS. This decrease highlights SEGO’s ability to generate more consistent and reliable semantic relationships at higher FPS. Conversely, relation entropy, which quantifies the diversity of semantic relations within the scene graph, increased with frame rate and began to saturate around 2.35 at 30 FPS and beyond.

Detailed inspection of violation-entropy scatter plots revealed distinct patterns across FPS conditions. At 30 FPS and below, the data points exhibited clear banded structures in the violation-entropy space, indicating that SEGO produces consistent cognitive scene graphs with stable, deterministic relation structures. In contrast, at 60 FPS, the scatter plot displayed a more dispersed pattern, suggesting increased variability due to higher frame rates, without corresponding improvements in semantic quality. This phenomenon further reinforces the observed saturation point near 30 FPS, where the system’s semantic graph stability and mapping efficiency were maximized.

\begin{figure}[!t]
\centering
\subfloat[30 FPS and below]{\includegraphics[width=0.48\linewidth]{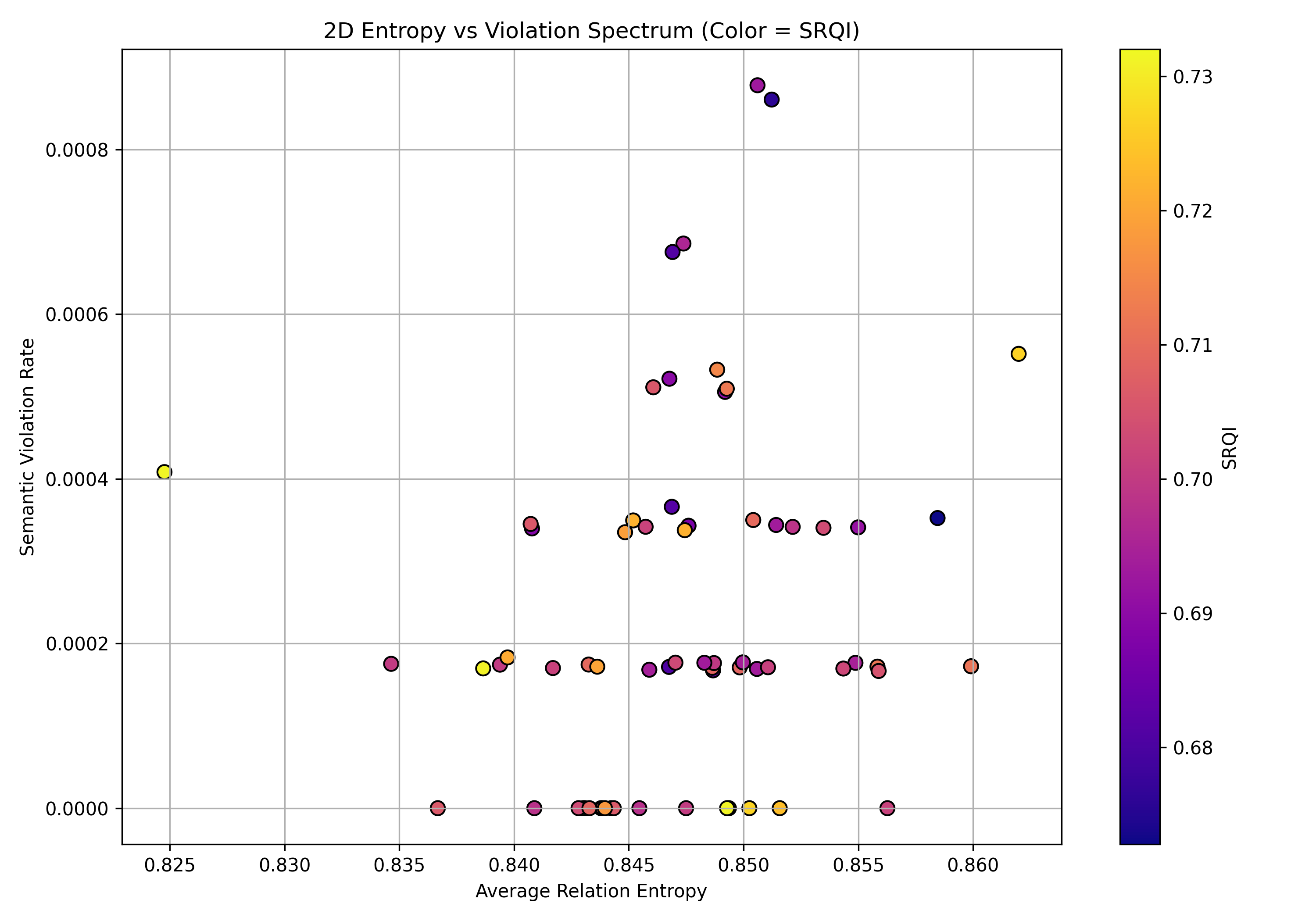}}
\hfil
\subfloat[60 FPS]{\includegraphics[width=0.48\linewidth]{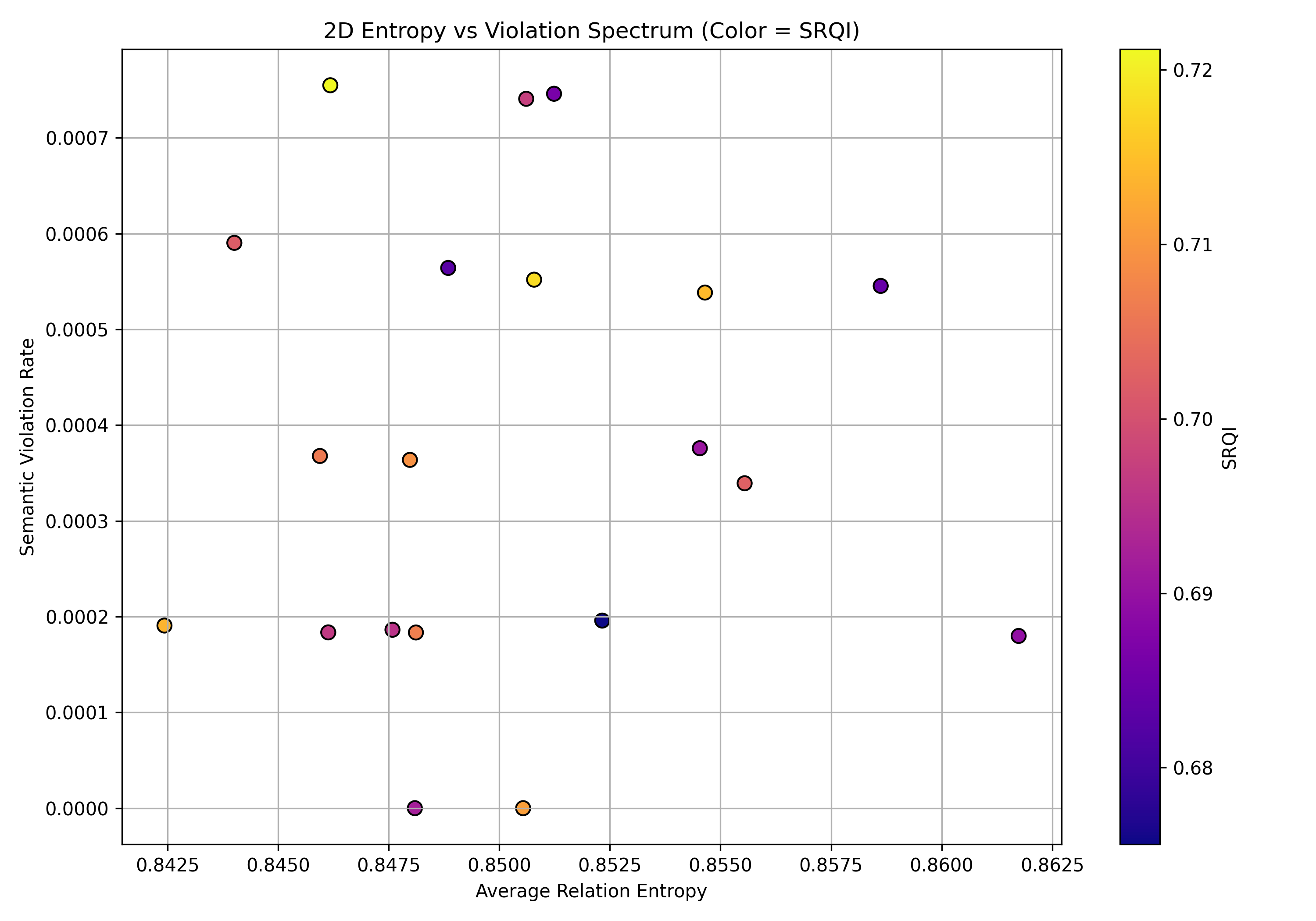}}
\caption{
Semantic violation rate vs relation entropy at different FPS settings.  
(a) At 30 FPS and below, SEGO generates stable cognitive scene graphs, visible as layered/banded data point patterns.  
(b) At 60 FPS, increased micro-variability and perceptual redundancy result in dispersed violation-entropy distributions without further semantic quality improvement.
}
\label{fig:violation_entropy}
\end{figure}

\subsubsection{Scene Graph Structural Complexity}
The structural complexity of the generated scene graphs was quantified in terms of node and edge counts, average degree, and clustering coefficient. These metrics revealed that as FPS increased, the complexity of the scene graph also increased. However, beyond 30 FPS, the rate of increase in these metrics diminished, indicating that additional frame rate improvements no longer resulted in proportionate gains in graph complexity. This supports the findings that frame rate beyond 30 FPS does not significantly enhance the richness of the cognitive scene graph, reinforcing the identified saturation point.

\subsubsection{Computational Cost}
\begin{figure}[!t]
\centering
\includegraphics[width=\linewidth]{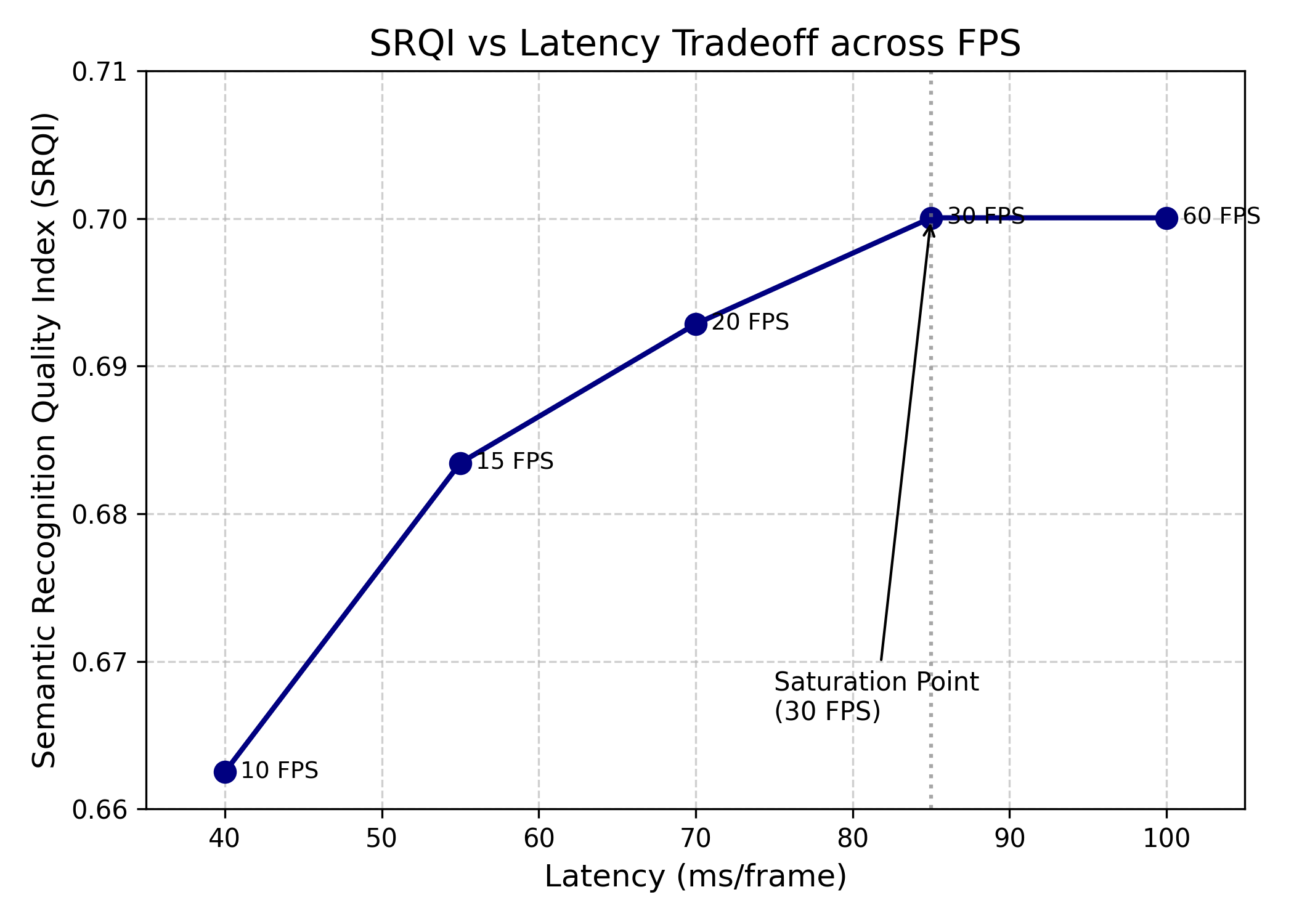}
\caption{SRQI vs latency across FPS settings, highlighting tradeoff curve.}
\label{fig:srqi_latency_tradeoff}
\end{figure}

The computational cost, including latency and resource usage, was also evaluated. Latency decreased slightly as FPS increased; however, the marginal gain in SRQI per millisecond of added latency beyond 30 FPS was negligible. This tradeoff between SRQI improvement and latency is illustrated in \textbf{Fig.~\ref{fig:srqi_latency_tradeoff}}. SEGO’s design achieves an optimal balance between performance and computational efficiency at 30 FPS, making it a viable solution for real-time applications.

\subsection{Qualitative Results}

\subsubsection{Example Scene Graphs}
\begin{figure}[!t]
\centering
\includegraphics[width=\linewidth]{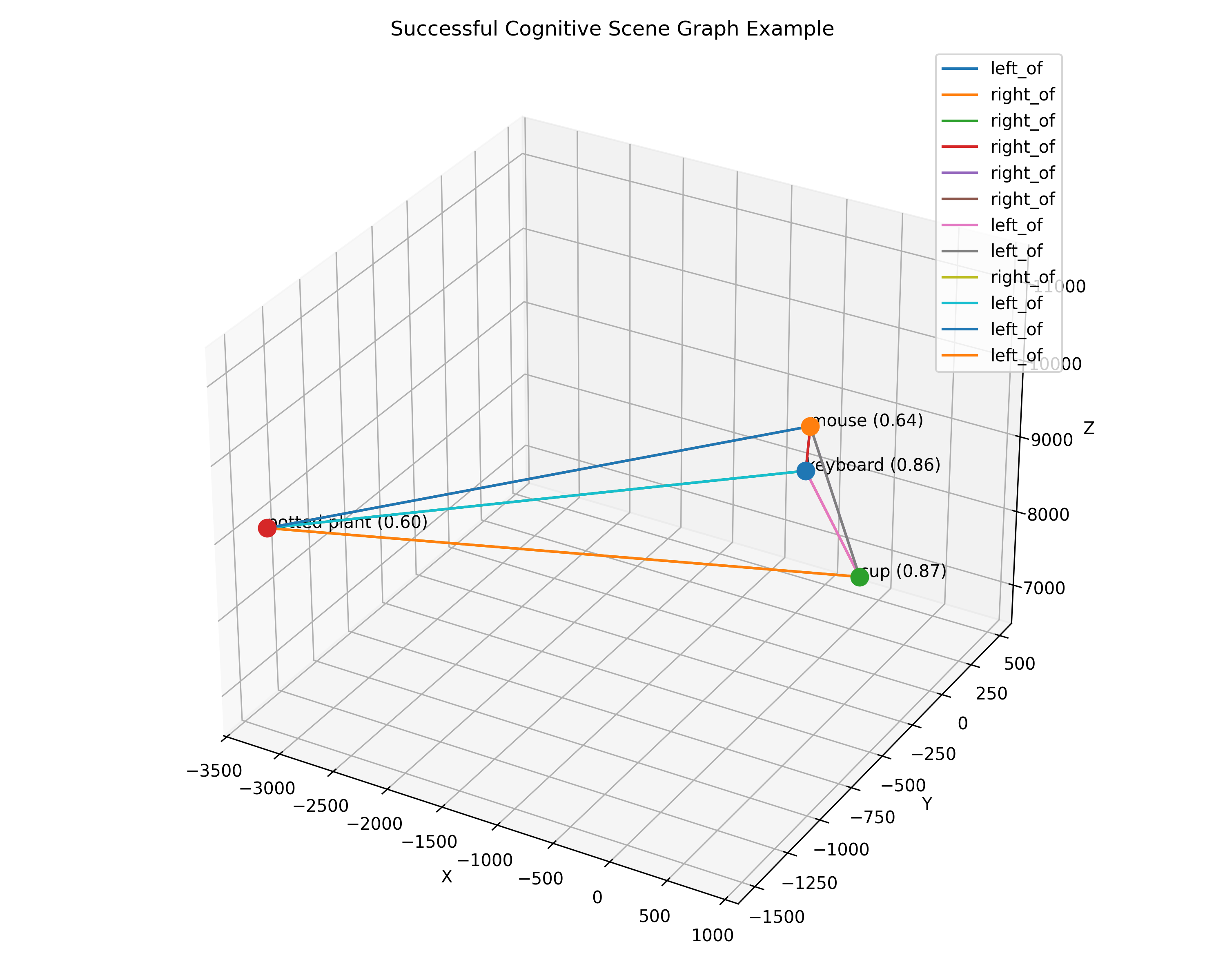}
\caption{Example cognitive scene graphs at varying FPS.}
\label{fig:scene_graphs}
\end{figure}

The qualitative evaluation of SEGO’s cognitive scene graphs at different FPS conditions is depicted in \textbf{Fig.~\ref{fig:scene_graphs}}. As the FPS increased from 10 to 60, the scene graphs became more densely populated, with better semantic coherence and greater node-edge connectivity. However, the gains at 60 FPS were marginal, reinforcing the observation that higher FPS beyond 30 offers limited improvements in terms of graph quality.

\subsubsection{Explainability Trace Examples}
\begin{figure}[!t]
\centering
\includegraphics[width=0.6\linewidth]{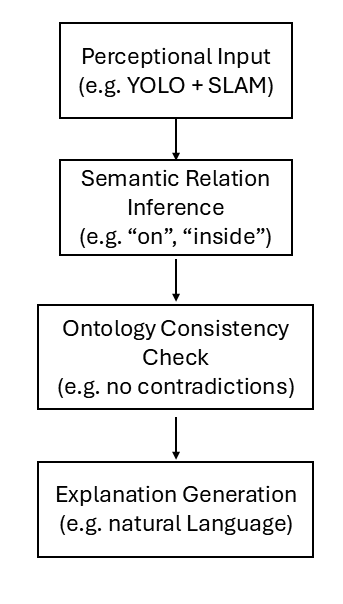}
\caption{Explainability reasoning flow example.}
\label{fig:explainability_flow}
\end{figure}

SEGO’s ability to generate transparent and explainable decision-making is crucial for human-robot interaction. The explanation chains generated by SEGO link the perceptual data to the reasoning process, allowing human collaborators to understand why a particular decision was made. Example explanation chains include:
\begin{itemize}
    \item \textit{“The bottle is classified as on the table because its centroid projects within the table’s area...”}
    \item \textit{“The cup is inside the cabinet because its volume fully intersects the cabinet’s interior.”}
\end{itemize}
These traces are visualized in \textbf{Fig.~\ref{fig:explainability_flow}}, demonstrating SEGO’s capability for real-time, understandable explanations.

\subsubsection{Failure and Edge Cases}
\begin{figure}[!t]
\centering
\includegraphics[width=\linewidth]{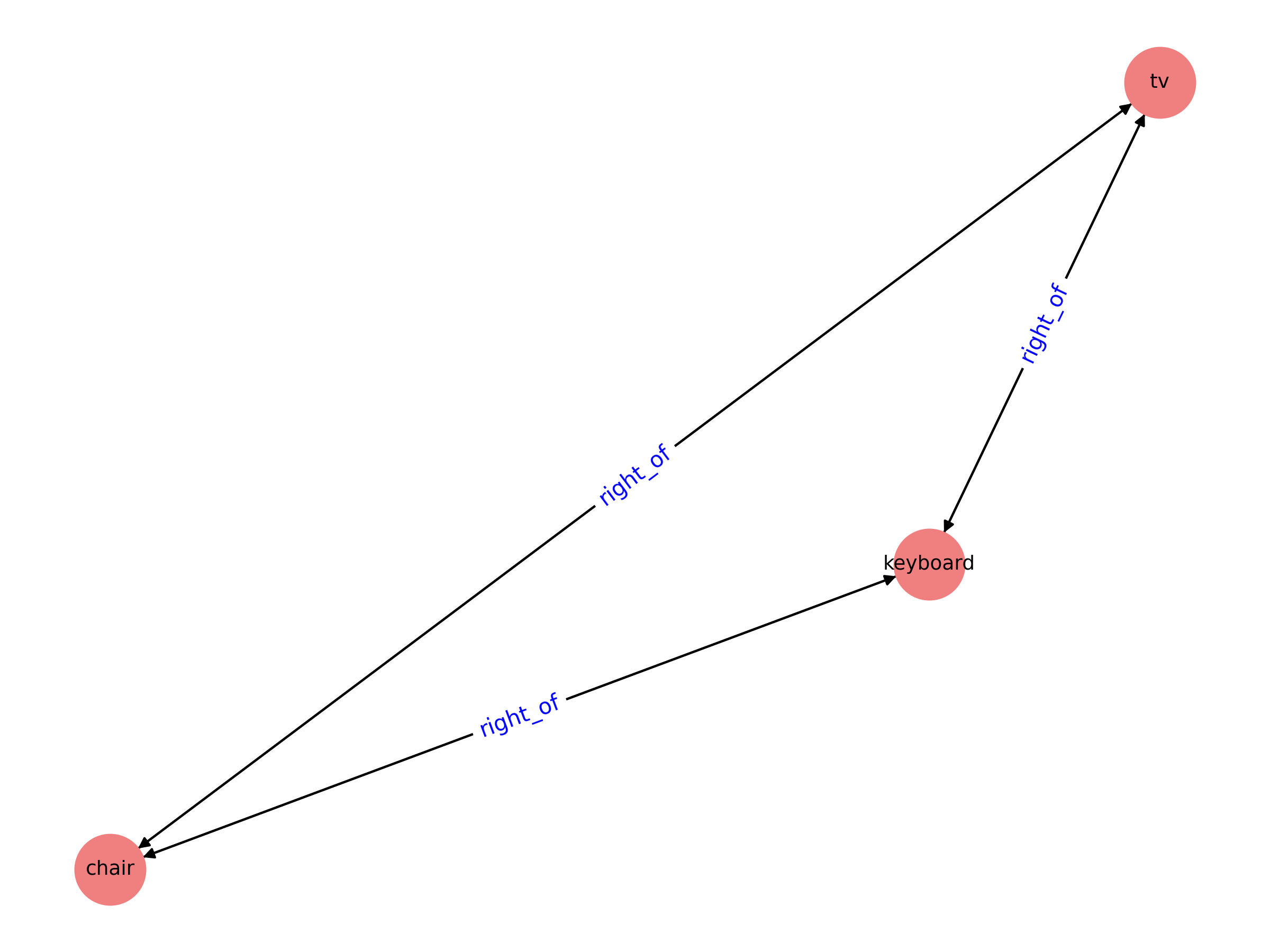}
\caption{Selected failure cases at low FPS.}
\label{fig:failures}
\end{figure}

Failure cases, such as occlusion, depth noise, and stale pose data, were observed predominantly at lower frame rates. At 10 FPS, tracking discontinuities and positional drift led to incorrect semantic relations in the generated scene graph. \textbf{Fig.~\ref{fig:failures}} highlights several of these failure cases, providing visual insight into how low FPS conditions adversely affect SEGO’s mapping and reasoning capabilities.

\subsection{FPS Saturation Analysis}
\begin{figure}[!t]
\centering
\includegraphics[width=\linewidth]{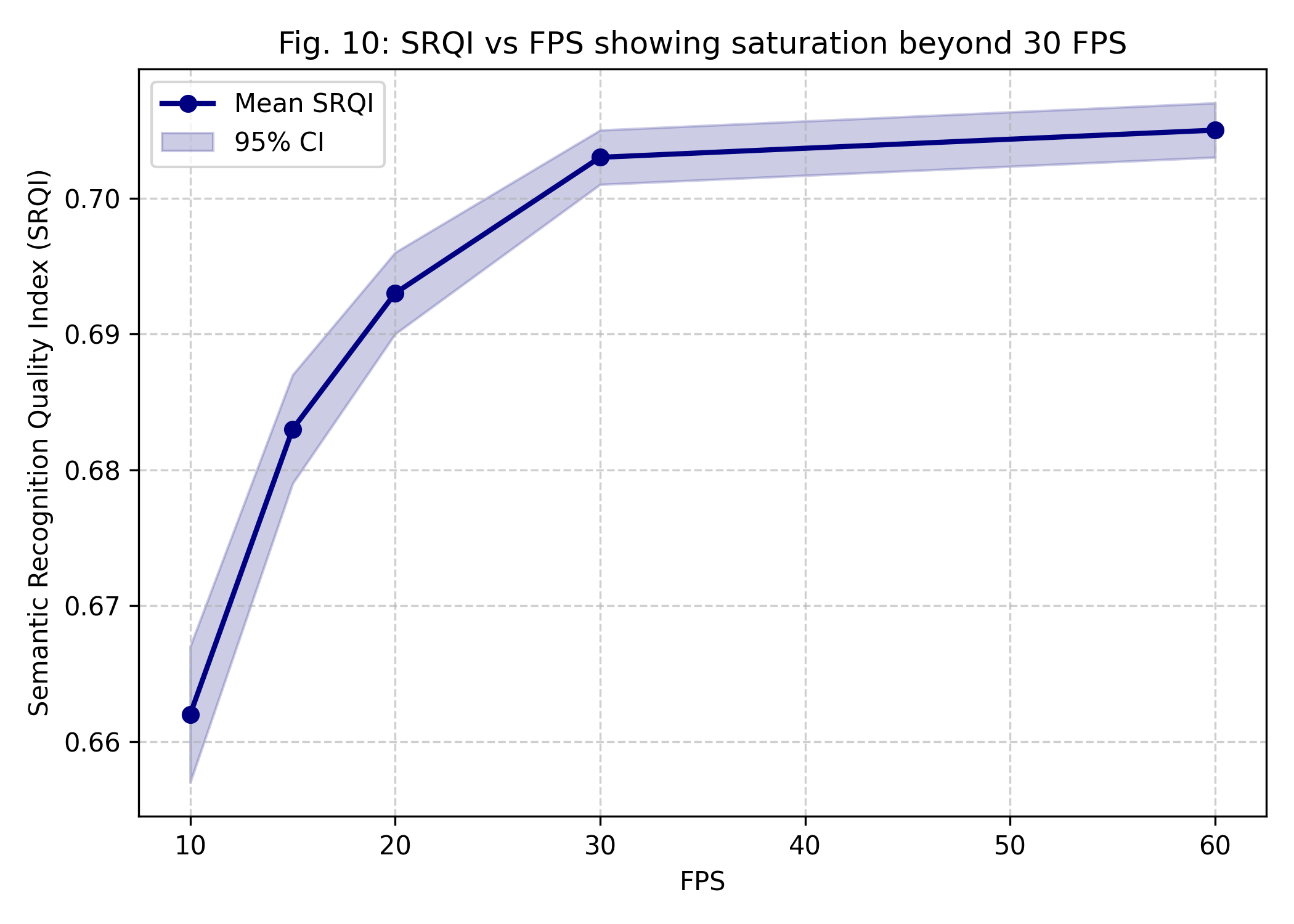}
\caption{SRQI vs FPS showing saturation beyond 30 FPS.}
\label{fig:fps_saturation}
\end{figure}

As previously discussed, SEGO exhibits a saturation effect at 30 FPS, where improvements in SRQI, violation rate, and entropy plateau. This saturation effect is important for understanding the trade-off between computational resources and performance. Beyond 30 FPS, the benefits are marginal, and the system operates optimally at this frame rate. The SRQI vs FPS curve shown in \textbf{Fig.~\ref{fig:fps_saturation}} further illustrates this saturation.

\subsection{Integrated Summary}
\begin{figure}[!t]
\centering
\includegraphics[width=\linewidth]{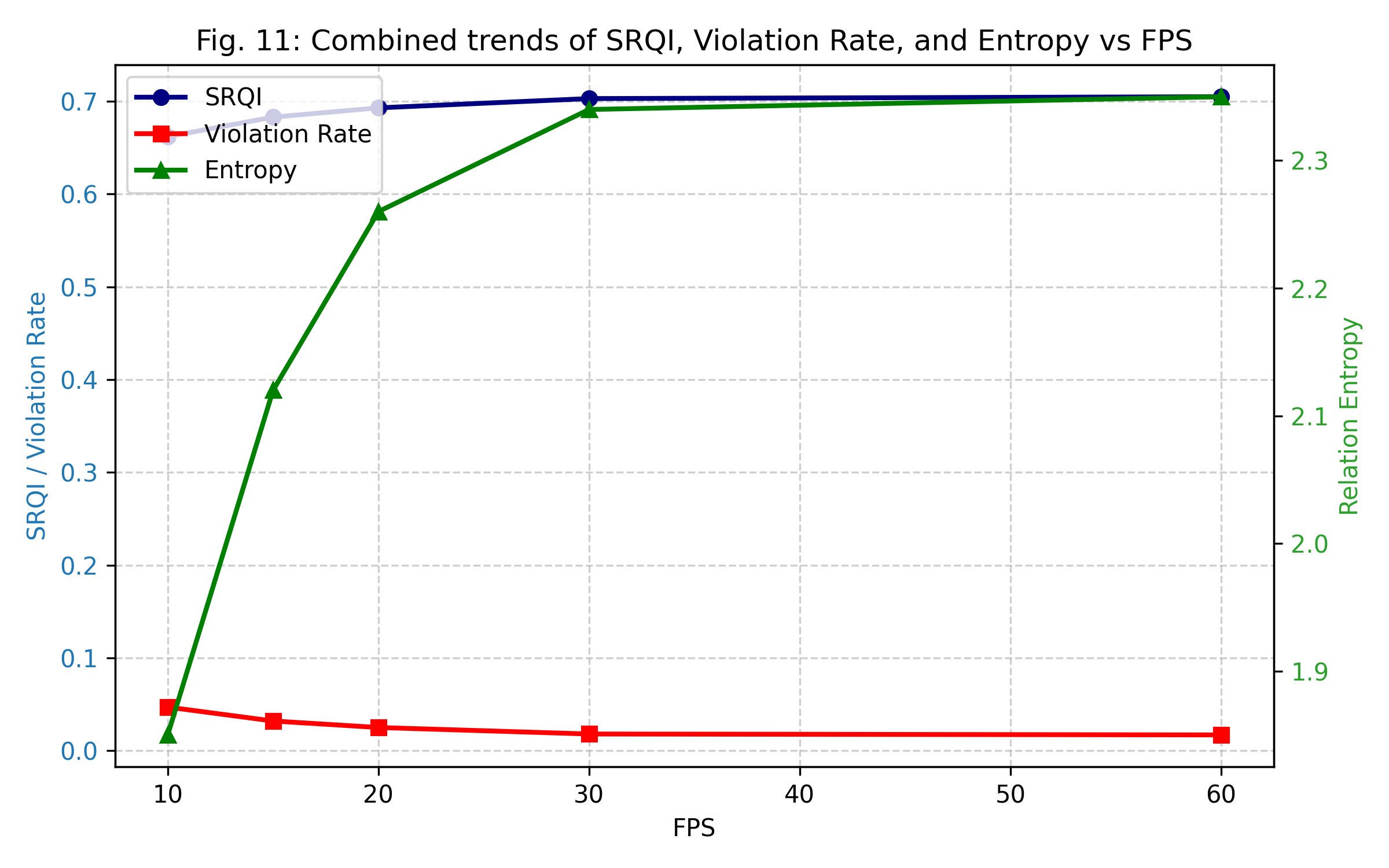}
\caption{Combined trends of SRQI, Violation Rate, and Entropy vs FPS.}
\label{fig:summary_trends}
\end{figure}

Finally, Table~\ref{tab:summary_metrics} consolidates the critical performance metrics across different FPS settings. Fig.~\ref{fig:summary_trends} overlays key trends, including SRQI, violation rate, and relation entropy, highlighting the most significant improvements at 30 FPS and beyond.

\begin{table}[!t]
\caption{Integrated Summary of Key Metrics}
\label{tab:summary_metrics}
\centering
\begin{tabular}{|c|c|c|c|}
\hline
FPS & SRQI & Violation Rate & Entropy \\
\hline
10 & 0.662 & 0.047 & 1.85 \\
15 & 0.683 & 0.032 & 2.12 \\
20 & 0.693 & 0.025 & 2.26 \\
30 & 0.703 & 0.018 & 2.34 \\
60 & 0.705 & 0.017 & 2.35 \\
\hline
\end{tabular}
\end{table}

\section{Discussion}

\subsection{Interpretation of FPS Saturation}
The experiments revealed a pronounced saturation effect in semantic mapping quality near 30 FPS. The Semantic Recognition Quality Index (SRQI) increased significantly from 0.662 at 10 FPS to 0.703 at 30 FPS, but demonstrated negligible improvement at 60 FPS (0.705). A similar trend was observed for violation rate, which decreased sharply to 0.018 at 30 FPS, with further reductions being minimal beyond this point. Relation entropy, indicative of relational diversity within the cognitive scene graph, rose from 1.85 at 10 FPS to 2.34 at 30 FPS, plateauing thereafter.  

These trends provide strong evidence that SEGO’s architecture fully exploits perceptual data at 30 FPS, where both semantic richness and logical coherence are maximized without incurring additional computational overhead. The saturation point not only marks an inflection in system efficiency but also serves as an empirical validation of the design hypothesis that a frame-rate-aware cognitive mapping pipeline can achieve human-aligned semantic quality while minimizing resource usage. This critical tradeoff, highlighted in \textbf{Fig.~\ref{fig:fps_saturation}}, underscores SEGO's capability to balance semantic fidelity with real-time operational demands.

\subsection{Comparison to Human Perceptual FPS}
The identified saturation point closely aligns with the perceptual integration limits of the human visual system, typically reported between 24–30 FPS in visual psychophysics and cognitive neuroscience literature \cite{watson1986temporal}. This alignment is not merely coincidental; it reflects SEGO's capacity to synchronize its cognitive map updates with temporal rhythms that are intuitive and natural for human collaborators. By operating within this perceptual sweet spot, SEGO facilitates shared situational awareness, mutual predictability, and fluid interaction in human-robot teams.

Moreover, this alignment has practical implications for system design. Operating beyond 30 FPS offers negligible semantic benefit but imposes disproportionate computational cost, particularly for embedded or mobile robotic platforms where processing power and energy reserves are constrained. The ability to cap frame rates intelligently while preserving semantic performance opens pathways to more sustainable and scalable robotic deployments.

\subsection{Implications for Explainability and HRI}
A defining feature of SEGO is its intrinsic support for explainability through reasoning traceability. The cognitive scene graph $\mathcal{G}(t) = (V_t, E_t)$ not only represents the spatial and semantic configuration of the environment but also encodes the provenance of each node and relation:
\begin{equation}
\mathcal{E}: (V_t, E_t) \mapsto R_t
\end{equation}
where $R_t$ denotes a reasoning trace comprising perceptual evidence and ontological validation steps.

This capability ensures that SEGO’s decisions are not opaque; rather, they are transparent and justifiable, which is critical for establishing trust, accountability, and predictability in collaborative scenarios. Such transparency supports human operators in understanding the robot’s decision-making process, debugging unexpected behavior, and aligning human-robot plans.  
The flow of reasoning from perception to explanation is conceptually summarized in Fig.~\ref{fig:explainability_flow}, offering a blueprint for integrating cognitive transparency into robotic architectures.

\subsection{Limitations and Challenges}
While SEGO demonstrates robust performance, several limitations highlight avenues for future research:
\begin{itemize}
    \item \textbf{Sensitivity to perception errors:} SEGO’s performance degrades in environments with significant occlusion, dynamic clutter, or depth noise, occasionally resulting in erroneous or spurious relations in the cognitive graph.
    \item \textbf{Low-FPS vulnerabilities:} At frame rates below 15 FPS, the system exhibited increased positional drift, tracking discontinuities, and relational instability, indicating that temporal resolution below a critical threshold undermines cognitive coherence.
    \item \textbf{Scalability under high complexity:} As scene graph size and relational density increased, the reasoning engine experienced latency, stressing real-time guarantees and highlighting the need for scalable reasoning strategies.
\end{itemize}

Future work will focus on addressing these challenges by:
\begin{itemize}
    \item Incorporating multi-view depth fusion and learning-based depth completion to enhance perceptual robustness.
    \item Developing hierarchical and incremental reasoning frameworks that enable scalable, low-latency consistency validation.
    \item Exploring probabilistic relational models and uncertainty-aware reasoning to gracefully manage ambiguity and partial knowledge in dynamic environments.
\end{itemize}

\subsection{Design Validation and Broader Impact}
The collective findings validate SEGO’s architectural principles and engineering contributions:
\begin{itemize}
    \item Seamless fusion of SLAM-based geometric localization, deep-learning-based detection, and ontological reasoning, enabling principled cognitive map construction.
    \item Real-time generation of cognitive scene graphs with embedded explainability, supporting transparency and human-aligned situational awareness.
    \item Frame-rate-aware design, achieving semantic saturation at 30 FPS while optimizing computational and energy efficiency for deployment on practical robotic platforms.
\end{itemize}

These attributes position SEGO not merely as a technical advance in cognitive mapping, but as a foundational architecture for future robotic systems that aspire to operate transparently, efficiently, and collaboratively in complex, human-centered environments. Its design philosophy offers a blueprint for the next generation of cognitive robots capable of reasoning, explaining, and cooperating at human-compatible levels of performance.

\section{Conclusion}

\subsection{Summary of Key Findings}
This study introduced SEGO, a comprehensive cognitive mapping framework that unifies perception, semantic reasoning, and explanation generation to construct dynamic, semantically coherent cognitive scene graphs in real time. SEGO demonstrated substantial improvements in semantic mapping quality as perceptual frame rate increased, with the Semantic Recognition Quality Index (SRQI) rising from 0.662 at 10 FPS to 0.703 at 30 FPS, and exhibiting saturation beyond this point. This empirically validated saturation point aligns with the known limits of human perceptual integration (24--30 FPS) \cite{watson1986temporal}, underscoring SEGO's potential for facilitating natural, intuitive human-robot collaboration.  

Furthermore, SEGO's integrated explainability mechanisms and ontology-based reasoning modules enable transparent, accountable, and predictable decision-making processes, which are essential for fostering trust and shared situational awareness in collaborative robotic systems.

\subsection{Unique Contributions of SEGO}
SEGO contributes several key innovations to the field of cognitive robotics:
\begin{itemize}
    \item A unified architecture that seamlessly integrates SLAM-based geometric localization, YOLOv5 + StrongSORT tracking, dynamic scene graph construction, and ontology-based reasoning for logical validation and consistency enforcement.
    \item A real-time explanation generation capability that provides perceptually grounded, traceable justifications for robot decisions, linking sensor data to reasoning pathways in a human-comprehensible form.
    \item The first quantitative framework for evaluating semantic mapping quality as a function of frame rate, introducing novel metrics including SRQI, violation rate, relation entropy, and structural complexity indicators.
\end{itemize}
Collectively, these contributions position SEGO as a principled, scalable, and transparent solution capable of advancing the state of the art in cognitive mapping for human-centered robotic systems.

\subsection{Implications for Cognitive Robotics and HRI}
The findings and design philosophy of SEGO have significant implications for the broader field of cognitive robotics and human-robot interaction (HRI):
\begin{itemize}
    \item By embedding explanation traceability and ontological validation within the cognitive mapping pipeline, SEGO provides a foundation for transparent and interpretable robotic behavior, addressing critical challenges in the deployment of autonomous systems in human environments.
    \item The alignment of SEGO’s semantic mapping dynamics with human perceptual rhythms supports shared situational awareness, natural joint decision-making, and fluid human-robot collaboration.
    \item The frame-rate-aware architecture ensures that SEGO achieves high semantic fidelity without incurring unnecessary computational overhead, a property that is particularly beneficial for resource-constrained platforms such as mobile service robots, aerial drones, and field-deployable autonomous agents.
\end{itemize}
These attributes collectively establish SEGO as a robust and adaptable cognitive architecture that can serve as a foundation for the next generation of collaborative, explainable, and efficient robotic systems.

\subsection{Future Work Directions}
Building on the foundation established in this study, several promising directions for future research emerge:
\begin{itemize}
    \item \textbf{Distributed cognitive mapping:} Extending SEGO to multi-robot systems to enable distributed cognitive scene graph construction and shared situational awareness across heterogeneous agents.
    \item \textbf{Online learning and adaptation:} Incorporating mechanisms for dynamic ontology refinement and relational inference updates based on accumulated experience in evolving environments.
    \item \textbf{HRI-centric validation:} Conducting empirical user studies to assess SEGO’s explainability, transparency, and collaborative efficacy in real-world human-robot teaming scenarios.
    \item \textbf{Natural language and large language model (LLM) integration:} Enhancing SEGO’s interaction capabilities through the integration of LLMs to support context-aware, fluent natural language explanations and dialogue-based reasoning.
    \item \textbf{Scalable reasoning architectures:} Investigating hierarchical, incremental, and probabilistic reasoning frameworks to further improve the scalability and robustness of SEGO’s cognitive mapping pipeline in complex, unstructured environments.
\end{itemize}

Through these future efforts, SEGO can evolve into an even more versatile and powerful cognitive framework, further bridging the gap between autonomous robotic cognition and human-compatible, transparent decision-making in collaborative contexts.

\nocite{*}

\bibliographystyle{IEEEtran}
\bibliography{bibliography}

\end{document}